\documentclass{article}
\usepackage{spconf,amsmath,graphicx}
\usepackage{booktabs}
\usepackage[numbers,sort&compress]{natbib}

\newcommand{\tabincell}[2]{\begin{tabular}{@{}#1@{}}#2\end{tabular}}

\title{Multi-Stage Contrastive Regression for Action Quality Assessment}
\name{Qi An, Mengshi Qi, Huadong Ma}
\address{Beijing Key Laboratory of Intelligent Telecommunications Software and Multimedia\\
    Beijing University of Posts and Telecommunications, China\\
    \emph{\{anqi\_bupt,qms,mhd\}@bupt.edu.cn.}
    }

\begin{document}
%
\maketitle
\begin{abstract}
In recent years, there has been growing interest in the video-based action quality assessment~(AQA). Most existing methods typically solve AQA problem by considering the entire video yet overlooking the inherent stage-level characteristics of actions. To address this issue, we design a novel Multi-stage Contrastive Regression~(MCoRe) framework for the AQA task. This approach allows us to efficiently extract spatial-temporal information, while simultaneously reducing computational costs by segmenting the input video into multiple stages or procedures. Inspired by the graph contrastive learning, we propose a new stage-wise contrastive learning loss function to enhance performance. As a result, MCoRe demonstrates the state-of-the-art result so far on the widely-adopted fine-grained AQA dataset. Our source code is available at \textit{https://github.com/Angel-1999/MCoRe}.
\end{abstract}
\section{Introduction}
\label{sec:intro}

\let\thefootnote\relax
\footnotetext{This work was supported by Funds for Innovation Research Group Project of NSFC under Grant 61921003, NSFC Project under Grant 62202063, and 111 Project under Grant B18008.}


Action quality assessment~(AQA) refers to the evaluation of action execution and scores estimation through performance analysis, which is one of the crucial techniques for comprehending athletes' performances in videos and has already drawn increasing attention in recent years. However, AQA is a more challenging task compared with action recognition, detection, or segmentation. This is due to the fact that an action can be identified based on limited images, while assessing motion performance need to understand the whole action across the entire sequence.

Most of existing AQA methods~\cite{parmar2019and,tang2020uncertainty,yu2021group,bertasius2017baller,li2022pairwise} only concentrate on how to extract features of the whole action in a video. Generally, these methods employ a universal backbone network to process the complete video sequence and capture deep features of the video to estimate the quality score. However, these traditional AQA methods do not consider the multi-stage characteristics of complex actions. 
Several research~\cite{xu2022finediving,zhang2023learning} found that aligning each stage of action video can lead to a more accurate analysis of the differences between actions. 
Considering the multi-stage characteristics of complex actions in sports, we propose a new method to analyze and evaluate videos under segmented stages for the AQA issue.



Traditional video feature extraction methods~\cite{carreira2017quo,9351755,8621027,8954105,9052709} split the input video into clips with a fixed number of frames, then adopt a network of specified structures for global processing these clips or segments, which is not necessarily optimal for precise segmentation of competitive sports videos.
Therefore, we design a new 2D CNN-based method and incorporate the Gate Shift Module~\cite{sudhakaran2020gate} to extract the feature of each frame in the video, and then use the sequential model to learn a longer range of contextual information and divide the video into multiple steps or procedures. Furthermore, we propose a loss function based on graph contrastive learning~\cite{you2020graph,zhu2021graph} to improve segmentation performance.
To enhance the accuracy of stage segmentation, we generate positive and negative pairs by leveraging inter-video and intra-video information.

It is worthy to point out that using contrastive regression between action videos~\cite{yu2021group} can help models perceive small differences between similar videos. Inspired by that, we improved the video contrastive regression framework by comparing query and exemplar videos at corresponding stages, to enhance our understanding of their relative differences.








This paper makes the following contributions: (1) We propose a multi-stage contrastive regression framework for the procedure segmentation and action quality assessment. (2) We introduce a contrastive loss that compares the intra-video and inter-video stages information for the reasonable procedure segmentation. (3) We achieve state-of-the-art performance on the fine-grained AQA dataset~\cite{xu2022finediving}.

\begin{figure*}[t]
  \centering
  \includegraphics[width=\linewidth]{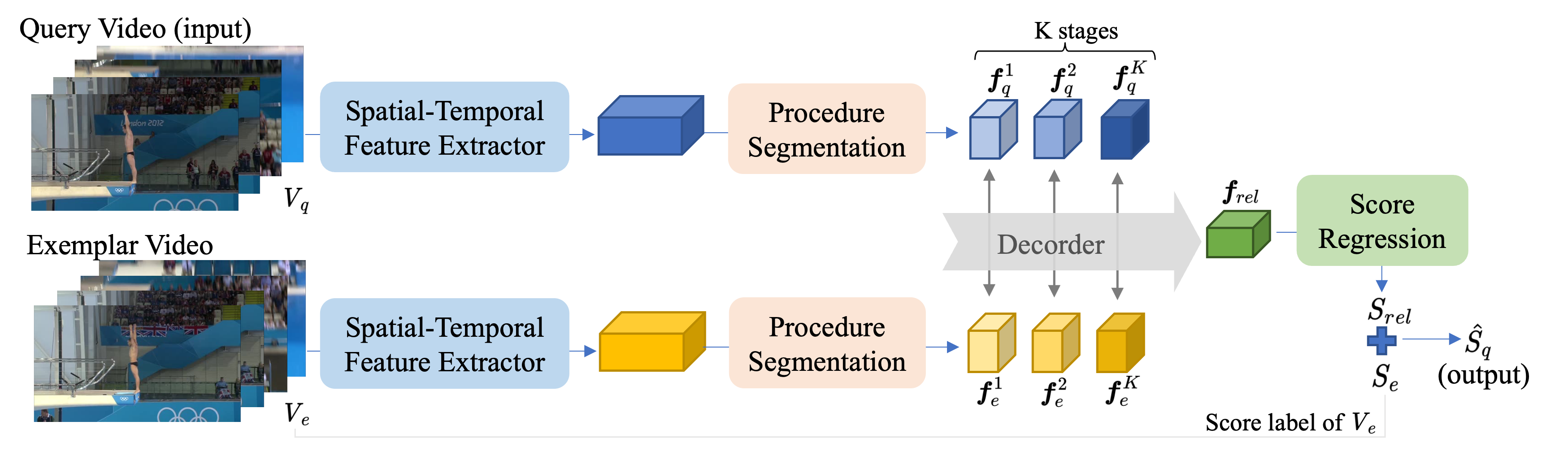}
  \caption{The overview of MCoRe Network. The input video to be tested is denoted as $\mathit{V_q}$, and $\mathit{V_e}$ is the selected exemplar from an existing dataset. Features are separately extracted from $\mathit{V_q}$ and $\mathit{V_e}$, and then divided into K stages. Considering the utilization of features from the same stages as a pair, we employ a decoder to obtain $\boldsymbol{f}_{rel}$ , which represents the difference between the two videos. Finally, relative scores are regressed, and added to score $\mathit{S_e}$, resulting in the predicted score $\hat{\mathit{S_q}}$. }
  \label{fig:NetWorkArchitecture}
\end{figure*}
\section{METHOD}
\label{sec:format}

In this section, we introduce our proposed Multi-stage Contrastive Regression (MCoRe) method in detail. We propose a novel framework to divide the pairwise videos into continuous stages, which are aligned and compared to regress the relative score. The overall network architecture is presented in Figure \ref{fig:NetWorkArchitecture}. 

\subsection{Overview}

\textbf{Problem Formulation.} Following~\cite{yu2021group}, we define the problem as follows: Given a sport video  $\mathit{V} = \{\mathit{v}_{i}\}_{i=1}^L $, where $\mathit{v}_{i}$ is the \textit{i}-th frame, and $L$ is the length of the video. $\mathit{V_q}$ represents the input query video. And the exemplar video has been sampled with the same diving type number (\emph{e.g.,} ``407C'' in rules of diving) as query, which means they have the same stage and content of the action. $\mathit{V_e}$ represents the exemplar video of the input video $\mathit{V_q}$. The score label of the video $\mathit{V_q}$ , $\mathit{V_e}$ are detenoted as $\mathit{S_q}$ , $\mathit{S_e}$ , respectively.
The predicted score $\hat{\mathit{S_q}}$ can be formulated as follows:
\begin{equation}
\hat{\mathit{S_q}} = \mathit{S_e} + \Upsilon (\mathit{V_q} ;\mathit{V_e}),
\end{equation}
where $\Upsilon(*;*)$ represents our multi-stage contrastive regression framework, and the result of $\Upsilon (\mathit{V_q} ;\mathit{V_e})$ is the predicted relative score.

Specifically, our proposed framework consists of a three-part network:~$\mathcal{F},\mathcal{D},\mathcal{R}$ represent the feature extractor network, procedure segmentation network and scoring network, respectively. The relative score can be regressed by computing:
\begin{equation}
\Upsilon (\mathit{V_q} ;\mathit{V_e}) = \mathcal{R}(\mathcal{D}(\mathcal{F}(\mathit{V_q}));\mathcal{D}(\mathcal{F}(\mathit{V_e}))).
\end{equation}
Assuming the input action video should be divided into \(K\) stages, we refer segmented features as \(\boldsymbol{f}_{q} =[\boldsymbol{f}_{q}^{1},\boldsymbol{f}_{q}^{2},...,\boldsymbol{f}_{q}^{K}] = \mathcal{D}(\mathcal{F}(\mathit{V_q}))\), and 
\(\boldsymbol{f}_{e} =[\boldsymbol{f}_{e}^{1},\boldsymbol{f}_{e}^{2},...,\boldsymbol{f}_{e}^{K}] = \mathcal{D}(\mathcal{F}(\mathit{V_e}))\).

\subsection{Spatial-Temporal Feature Extractor}


The initial phase of the framework involves the spatial-temporal feature extractor network, which aims to enhance the comprehension of the videos. 
Previous attempts in AQA typically involved extracting temporal features from non-overlapping video clips, with each feature consisting of consecutive frames, such as I3D~\cite{carreira2017quo} and S3D~\cite{xie2018rethinking}. However, these approaches exhibit limited precision when precise stage segmentation is required. To overcome this challenge, we have employed the improved RegNet-Y~\cite{radosavovic2020designing} architecture as the 2D CNN backbone. The feature extracted by the backbone is denoted as $ X\ =\mathcal{F}(\mathit{V})$.


This approach is motivated by the desire to encompass spatial image information from every frame, which is vital for precise procedure segmentation. However, 2D CNNs are unable to capture the spatial-temporal correlations between frames. In high-speed activities such as diving, motion blur may affect each frame in the video, making it important to obtain features that are more sensitive to motion change. Thus, we integrated Gate Shift Module~(GSM)~\cite{sudhakaran2020gate} into the RegNet-Y architecture. The inclusion of GSM enhances the ability of the feature extractor to discern temporal dynamics between consecutive frames, enabling it to capture the essential spatial-temporal variations for the AQA task.

\subsection{Procedure Segmentation}
In order to divide the features of the videos into stage-level representations, we employ a procedure segmentation network $\mathcal{D}$ here. It is designed to identify the \text{K-1} stage transitions frames in total. The ground truth stage transition label of the video is \([a_1,a_2,...a_L]\), and the \textit{i}-th frame stage label \(a_i\) is categorized into \text{K}  classes.
If the sub-action labels of the \textit{i}-th and \textit{i-1}-th frames are the same, it indicates that they belong to the same sub-action, and the transition label of the \textit{i}-th frame is 0. Otherwise, the transition label is 1.

To gather long-term temporal information, the network $\mathcal{D}$ is Bidirectional Gated Recurrent Unit (Bi-GRU) network~\cite{cho2014learning} and a fully connected layer followed by softmax operation. 
This network processes the features extracted by the backbone and predicts stage transition labels \([\hat{a}_1,\hat{a}_2,...,\hat{a}_L]\) for each frame.
To find transition positions, we select the frames with the highest probability as the prediction. 
We optimize the per-frame classification using cross-entropy loss. 
The loss function $\mathcal{L}_{ce}$ calculates the difference between the predicted stage transition moments and the ground truth, which can be formulated as the following:
\begin{equation}
\mathcal{L}_{ce} = \sum_{i=1}^K \text{CE}\left(a_i,\hat{a}_i\right).
\end{equation}

\subsection{Stage Contrastive Loss}
In order to enhance the precision of procedure segmentation in motion videos, we have developed a stage contrastive loss denoted as \(L_{\text{cont}}\). This loss utilizes a contrastive objective to distinguish features that belong to the same stage and different segmentation in two different videos.
Formally, we introduce the critic \(sim\left(\boldsymbol{f}_{q}^{i}, \boldsymbol{f}_{e}^{i}\right) \), where \({cos}\) denotes cosine similarity, and \({norm}\) denotes a normalization function used to enhance the critic's expressive capacity, as follows:
\begin{equation}
sim\left(\boldsymbol{f}_{q}^{i}, \boldsymbol{f}_{e}^{i}\right) = \cos{(norm(\boldsymbol{f}_{q}^{i}),norm(\boldsymbol{f}_{e}^{i}))} .
\end{equation}
During the computation of the contrastive loss, we treat the features from the same stage in query video and exemplar video as positive pairs, denoted as $\boldsymbol{\epsilon}(\boldsymbol{f}_{q}^{i},\boldsymbol{f}_{e}^{i})$. 
While negative pairs $\boldsymbol{\zeta}(\boldsymbol{f}_{q}^{i},\boldsymbol{f}_{e}^{i})$, can be divided into inter-video or intra-video pairs. Inter-video negative pair $(\boldsymbol{f}_{q}^{i},\boldsymbol{f}_{e}^{j})$ is the different stages across different videos, while intra-video pair $(\boldsymbol{f}_{q}^{i},\boldsymbol{f}_{q}^{j})$ is the different stages within the same video. 
The positive pairs $\boldsymbol{\epsilon}(\boldsymbol{f}_{q}^{i},\boldsymbol{f}_{e}^{i})$ and negative pairs $\boldsymbol{\zeta}(\boldsymbol{f}_{q}^{i},\boldsymbol{f}_{e}^{i})$ are formulated as:

\begin{equation}
\boldsymbol{\epsilon}(\boldsymbol{f}_{q}^{i},\boldsymbol{f}_{e}^{i}) = e^{sim\left(\boldsymbol{f}_{q}^{i}, \boldsymbol{f}_{e}^{i}\right) / \tau},
\end{equation}
\begin{equation}
\boldsymbol{\zeta}(\boldsymbol{f}_{q}^{i},\boldsymbol{f}_{e}^{i}) = 
\underbrace{\sum_{j=1;j \neq i}^K e^{sim\left(\boldsymbol{f}_{q}^{i}, \boldsymbol{f}_{e}^{j}\right) / \tau}}_{\text {inter-video negative pairs }}
+
\underbrace{\sum_{j=1;j \neq i}^K e^{sim\left(\boldsymbol{f}_{q}^{i}, \boldsymbol{f}_{q}^{j}\right) / \tau}}_{\text {intra-video negative pairs }},
\end{equation}
where $\tau$ denotes the temperature parameter and we set as 0.5 in practice.
We define the pairwise objective for pair  $\left(\boldsymbol{f}_{q}^{i},\boldsymbol{f}_{e}^{i}\right)$ as:
\begin{equation}
\ell\left(\boldsymbol{f}_{q}^{i},\boldsymbol{f}_{e}^{i}\right)= 
\log \frac{
\boldsymbol{\epsilon}(\boldsymbol{f}_{q}^{i},\boldsymbol{f}_{e}^{i})
}
{
\boldsymbol{\epsilon}(\boldsymbol{f}_{q}^{i},\boldsymbol{f}_{e}^{i})
+
\boldsymbol{\zeta}(\boldsymbol{f}_{q}^{i},\boldsymbol{f}_{e}^{i})
}.
\end{equation}
The stage contrastive Loss is defined as the average over all positive pairs, denoted as $\mathcal{L}_{cont}$, formulated as follows:
\begin{equation}
\mathcal{L}_{cont}=\frac{1}{2 K} \sum_{i=1}^K\left[\ell\left(\boldsymbol{f}_{q}^{i},\boldsymbol{f}_{e}^{i}\right)+\ell\left(\boldsymbol{f}_{e}^{i},\boldsymbol{f}_{q}^{i}\right)\right].
\end{equation}
\subsection{Scoring}
After dividing the video features, we uniformly fix the size of segmented video in temporal dimension by linear interpolation method. Then we obtain stage-level features with semantic and temporal correspondences for \(K\) stages.
We compute differences \(\boldsymbol{f}_{rel}^k\) between the \textit{k}-th stages in two videos using a decoder module with attention mechanism. Based on differences \(\boldsymbol{f}_{rel}=[\boldsymbol{f}_{rel}^1,\boldsymbol{f}_{rel}^2,...,\boldsymbol{f}_{rel}^K]\), we perform regression to obtain the relative score \(\mathit{S_{rel}}\), and predict the score of the query video \(\hat{\mathit{S_q}}\) as the following:
\begin{equation}
\boldsymbol{f}_{rel} = \left[\textit{Dec}(\boldsymbol{f}_{q}^{1},\boldsymbol{f}_{e}^{1})),...,\textit{Dec}(\boldsymbol{f}_{q}^{K},\boldsymbol{f}_{e}^{K})\right] ,
\end{equation}
\begin{equation}
\hat{\mathit{S_q}} = \mathit{S_e} + \mathit{S_{rel}} = \mathit{S_e} + \textit{Reg}\left(\boldsymbol{f}_{rel} \right) .
\end{equation}
The \(\textit{Dec}\) represents a decoder composed of multiple attention blocks. In each block, we employ Multi-head Cross-Attention~\cite{vaswani2017attention} to calculate the differences between \(\boldsymbol{f}_{q}^{k}\) and \(\boldsymbol{f}_{e}^{k}\). In addition, we incorporate residual connections and full-connected layers for enhanced modeling. The \textit{Reg} represents the MLP blocks. 

To evaluate the accuracy of score prediction in the AQA problem, we utilize the Mean Squared Error (MSE) as a metric. The MSE calculates the squared difference between the predicted scores and the ground truth values as the following:
\begin{equation}
\mathcal{L}_{aqa} = \text{MSE}\left(\mathit{S_q},\hat{\mathit{S_q}}\right) .
\end{equation}

\subsection{Optimization and Inference}
\textbf{Optimization.}
For each video pair in the training data with stage transition label \([a_1,a_2,...a_L]\) and score label $\mathit{S_q}$, the objective function for our task can be written as:
\begin{equation}
\mathcal{L}=\mathcal{L}_{aqa}+\mathcal{L}_{ce}+\mathcal{L}_{cont} .
\end{equation}
\textbf{Inference.}
We consider the input test video \(\mathit{V_{test}}\) as the query and extract \(P\) exemplar videos from the training. These exemplar videos are then paired with the query to form pairs \( (\mathit{V_{test}},\mathit{V_{e_p}})\). 
Using the trained MCoRe model, we assess action quality by employing stage alignment and contrastive regression techniques, which can be formulated as:
\begin{equation}
\hat{\mathit{S}}_{test} = \frac{1}{P}  \sum_{p=1}^P \left(  \mathit{S_{e_p}} + \Upsilon (\mathit{V_{test}} ;\mathit{V_{e_p}}) \right) .
\end{equation}

\section{EXPERIMENTS}
\subsection{Datasets and Metric}
\textbf{FineDiving Dataset.} Xu et al.~\cite{xu2022finediving} introduced a fine-grained dataset for action quality assessment, consisting of a total of 3000 diving clips. This dataset includes corresponding action labels (52 categories) and sub-action labels for each frame (29 categories). 
Compared with traditional AQA datasets, the FineDiving Dataset provides fine-grained classes and temporal boundaries for stages. 
\\
\textbf{Evaluation Metric.} 
To compare with existing state-of-the-art AQA methods, we use Spearman's rank correlation (SRCC) to quantify the rank correlation between the true scores and predicted scores. A higher SRCC indicates a stronger scoring capability of the model.
We also employ Relative l2-distance (R-l2) as an evaluation. 
Lower R-l2 indicates better performance of the approach.
The computation of SRCC and R-l2 metrics in detail can see in the paper~\cite{yu2021group}.

To assess the accuracy of stage segmentation, we compute the average intersection over union (AIoU) between two stage bounding boxes. And we determine the correctness of each prediction if the IoU is greater than a certain threshold \(d\), denoted as AIoU@\(d\).
\[
    \mathrm{AIoU} @ d=\frac{1}{N} \sum_{i=1}^N \mathcal{J}\left(\operatorname{IoU}_i \geq d\right) ,
\]
where if the judgment is true, \(\mathcal{J}=1\), otherwise, \(\mathcal{J}=0\). 
And a higher AIoU@\(d\) value indicates a better procedure segmentation performance. In this work, we primarily use AIoU@0.5 and AIoU@0.75 as the main metrics. 
\\
\textbf{Implementation Details.} 
We initialized the learning rates for the T as \(10^{-3}\) with the Adam optimizer in a cosine annealing schedule. 
During pre-processing, we extracted 96 frames from each raw video. 
We set the number of stage transitions \(K\) and exemplar videos \(P\) as 2 and 10, respectively.



\begin{table}[t]
  \caption{Comparison results with the state-of-the-art methods on FineDiving Dataset. Best results are in bold.}
  \begin{tabular}{cccc}
    \toprule
    Model & SRCC & R-l2(*100) & AIoU@0.5/0.75\\
    \midrule
    USDL~\cite{tang2020uncertainty} & 0.8913 & 0.3822 & - \\
    MUSDL~\cite{tang2020uncertainty} & 0.8978& 0.3704 & - \\
    CoRe~\cite{yu2021group} & 0.9061 & 0.3615 & - \\
    \midrule
    TSA~\cite{xu2022finediving} & 0.9203 & 0.3420 & 82.51 / 34.31\\
    \textbf{MCoRe} & \textbf{0.9232} & \textbf{0.3265
} & \textbf{98.26 / 79.17} \\
    \bottomrule
  \end{tabular}
  \label{methods}
  \vspace{-4mm}
\end{table}

\begin{table}[t]
  \caption{Operations comparison on FineDiving Dataset. }
  \begin{tabular}{cccc}
    \toprule
    Model & FLOPs & Params & Pretrained \\
    \midrule
    TSA~\cite{xu2022finediving} & 1008.7 G & 12.6 M & Imagenet \\
    \midrule
    \textbf{MCoRe} & \textbf{45.5 G} & \textbf{3.1 M} & -  \\
    MCoRe (w/o gsm) & 44.3 G & 3.0 M & -  \\
    \bottomrule
  \end{tabular}
  \label{operations}
   \vspace{-2mm}
\end{table}

\begin{table}[htb]
  \caption{Ablation studies of MCoRe.}
  \begin{tabular}{cccc}
    \toprule
    Model & SRCC & R-l2(*100) & AIoU@0.5/0.75\\
    \midrule
    \textbf{MCoRe} & \textbf{0.9232} & \textbf{0.3265} & \textbf{98.26 / 79.17} \\
    \tabincell{c}{MCoRe \\ (w/o $\mathcal{L}_{cont}$)} & 0.9206 & 0.3368 & 83.24 / 37.24 \\
    \bottomrule
  \end{tabular}
  \label{ablation}
   \vspace{-2mm}
\end{table}

\subsection{Results and Analysis}

As presented in Table \ref{methods}, MCoRe outperforms other methods across all metrics. 
Specifically, in terms of score evaluation, MCoRe achieves an SRCC of 0.9232 and an R-l2 (*100) of 0.3265.
We retrained USDL~\cite{tang2020uncertainty}, MUSDL~\cite{tang2020uncertainty}, and CoRe~\cite{yu2021group} on FineDiving without procedure segmentation, and the results were inferior to our approach. 
Moreover, MCoRe exhibits remarkable capabilities in procedure segmentation, attaining AIoU@0.5 / 0.75 scores of 98.26 / 79.17, which far surpass the performance of TSA~\cite{xu2022finediving}. 

Table \ref{operations} demonstrates that our approach is remarkably lightweight, primarily due to the integration of a 2D CNN with GSM as the backbone of our model.
MCoRe achieves comparable performance to state-of-the-art methods while requiring 22× fewer multiply-add operations (FLOPs) and 4× fewer parameters than TSA~\cite{xu2022finediving}. 
Moreover, there is no requirement for pretraining on larger-scale datasets beforehand, facilitating faster convergence.

An ablation study was conducted on MCoRe, and the results are presented in Table \ref{ablation}. Excluding \(\mathcal{L}_{cont}\) leads to a reduction of 0.0026 and 0.103 in the SRCC metric and R-l2(*100) metric, respectively. 
This demonstrates the effectiveness of the stage contrastive loss \(\mathcal{L}_{cont}\) in the design of our model.

\begin{figure}[htb]
  \centering
  \includegraphics[width=\columnwidth]{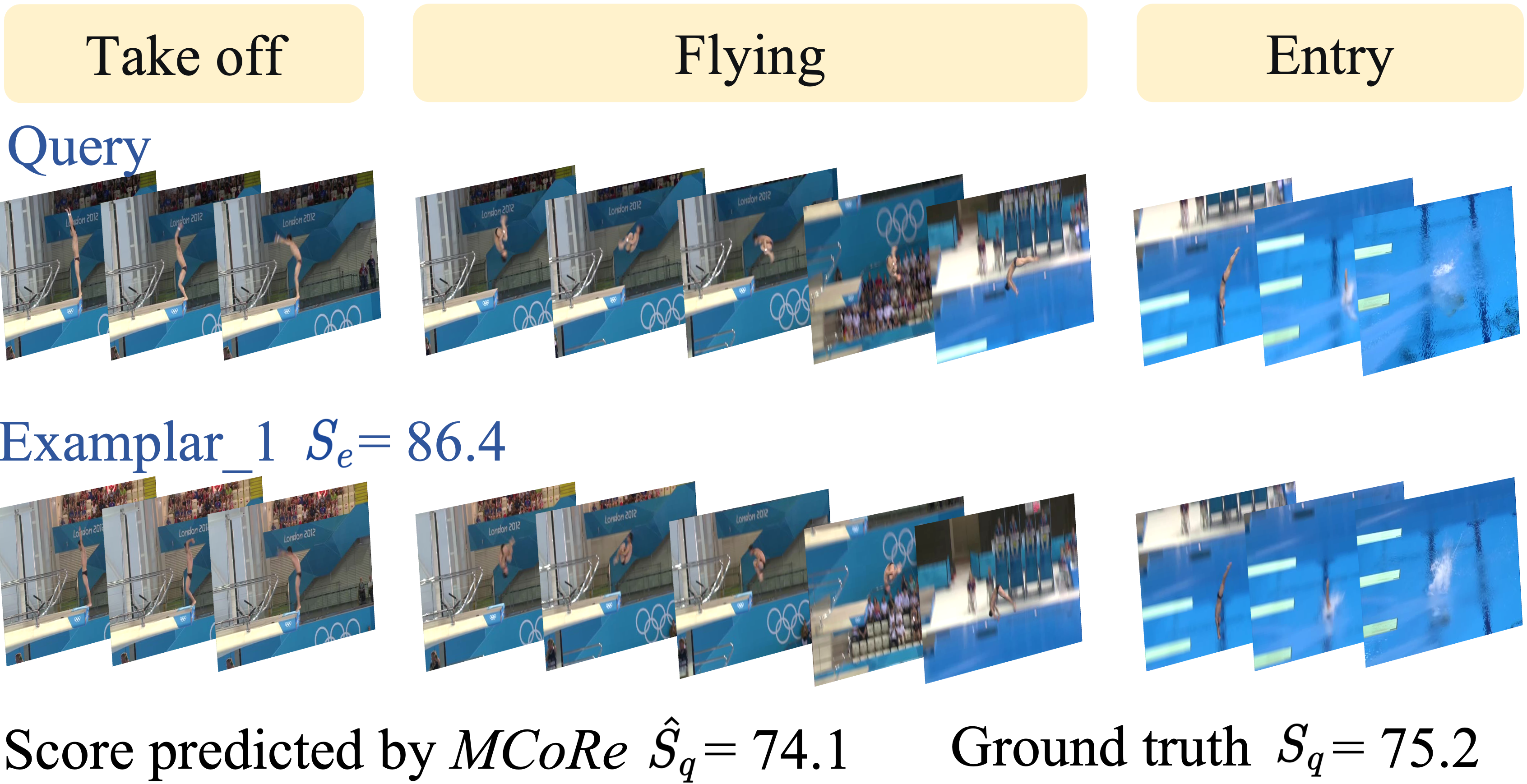}
  \caption{Visualized examples of the predicted results. Input a test video as query and randomly select videos from dataset with the same diving type number (e.g.,``407C'') as exemplars. }
  \label{fig:predicted}
  \vspace{-3mm}
\end{figure}

As shown in Figure \ref{fig:predicted}, the qualitative results clearly indicate that videos from the FineDiving dataset are segmented into three stages,  which stand for take off, flying, and entry in diving competition and our method produces predicted scores that closely align with the ground truth.

\section{Conclusion}
\label{sec:typestyle}

In this paper, we proposed MCoRe for multi-stage action quality assessment. Through the well-designed architecture, our model can efficiently extract spatial-temporal information from videos and reduce inference costs with the stage-wise contrastive loss, leading to more accurate assessment results on the fine-grained AQA dataset. 


\bibliographystyle{IEEEbib}
\bibliography{refs}

\end{document}